\documentclass[runningheads]{llncs}

\usepackage[T1]{fontenc}
\usepackage{graphicx}
\usepackage{amsmath,amssymb}
\usepackage{booktabs}
\usepackage{url}
\usepackage{bbding} 

\newcommand{\corrauth}{\textsuperscript{(\Envelope)}}

\begin{document}

\title{STEC: Evidence Compression for Deep Search in Open-domain Multi-Hop QA}
\titlerunning{STEC: Evidence Compression for Open-domain Multi-Hop QA}

\author{
Xinkang Li\inst{1}
\and Rong Jiang\inst{1}\corrauth
\and Xin Song\inst{1}
\and Ye Wang\inst{2}
\and Yue Han\inst{1}
\and Changjian Li\inst{1}
}

\authorrunning{X. Li et al.}

\institute{
National University of Defense Technology, China\\
\email{
xinkangli@nudt.edu.cn,
jiangrong@nudt.edu.cn,
songxin@nudt.edu.cn,\\
hanyue@nudt.edu.cn,
lichangjian23@nudt.edu.cn
}
\and
Harbin Institute of Technology, Shenzhen, China\\
\email{wangye2020@hit.edu.cn}
}

\maketitle

\begin{abstract}
In open-domain multi-hop question answering (QA), LLM-based search agents offer a promising approach to knowledge-intensive QA by combining retrieval with reasoning. Existing methods mainly improve open-domain multi-hop QA through reasoning paradigms, retrieval interaction, and search strategy optimization. However, using multiple search trajectories introduces a challenging final answer selection problem. Different trajectories may support different candidates, and the retrieved information can be heterogeneous, redundant, incomplete, or conflicting. Directly comparing raw trajectories exposes the verifier to noisy and unaligned content, while comparing answer strings ignores the evidence supporting each candidate, making reliable final selection difficult. To address this challenge, we propose STEC, an evidence compression framework for final answer selection in multi-hop QA. STEC selects the final answer from the existing candidate set through two mechanisms: (1) Answer-Level Evidence Compression, which groups trajectories by normalized answer identity and converts each answer group into a candidate-specific evidence representation; and (2) Evidence-Guided Answer Verification, which compares these representations and selects the final answer from the candidate set. The design shifts final selection from raw trajectory comparison to candidate-level evidence comparison. We evaluate STEC on four open-domain multi-hop QA benchmarks against representative baselines. Experimental results show that STEC performs best overall among the compared methods, and ablation results provide evidence that answer-level evidence compression contributes to final answer selection.

\end{abstract}

\section{Introduction}

\begin{figure}[t]
\centering
\includegraphics[width=0.90\textwidth]{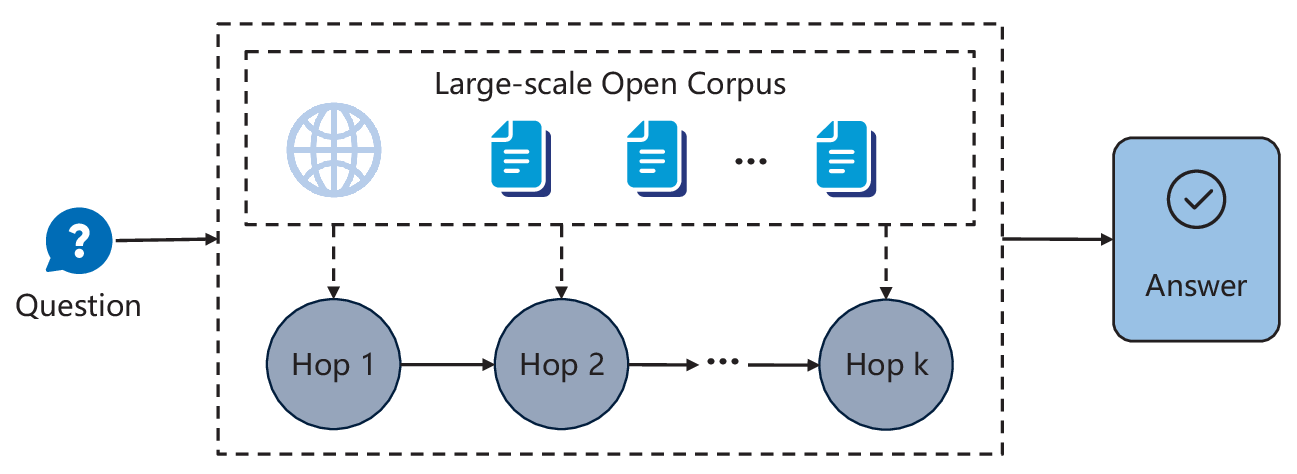}
\caption{Illustration of open-domain multi-hop question answering. A system retrieves relevant information from a large-scale open corpus and reasons across multiple hops to derive the final answer.}
\label{fig:intro_task}
\end{figure}

Open-domain multi-hop question answering is a challenging knowledge-intensive task. A system needs to retrieve relevant information from large-scale open corpora and integrate it across entities, relations, and events to support coherent multi-step reasoning. Unlike single-hop question answering, a multi-hop question usually cannot be answered with a single retrieved passage. The model must combine information from multiple sources through reliable multi-step reasoning, and retrieval omissions, information noise, or intermediate reasoning errors can all lead to an incorrect final answer. Fig.~\ref{fig:intro_task} illustrates this task setting.

Traditional retrieval-augmented generation provides a basic paradigm for knowledge-intensive question answering. The model first retrieves external context and then generates an answer based on the retrieved results, which mitigates knowledge gaps and factual errors that arise when LLMs rely only on parametric knowledge~\cite{ref1}. However, open-domain multi-hop question answering often involves entities, relations, and constraints that emerge gradually during reasoning. As reasoning proceeds, the model may need to resolve new intermediate entities, compare different facts, or verify conditions introduced by previous reasoning steps. A single retrieve-then-generate pipeline cannot continuously update retrieval targets according to the intermediate reasoning state. Once the initial retrieval fails to cover key evidence or introduces noisy information, subsequent reasoning is likely to rely on incomplete or even incorrect context.

To address this limitation, recent deep search methods incorporate retrieval into the step-by-step reasoning process. Representative methods such as Search-o1~\cite{ref5}, Search-R1~\cite{ref10}, and ReSearch~\cite{ref11} no longer treat external knowledge acquisition as a fixed preparation step before generation. Instead, they allow the model to issue search queries, read evidence, and use newly obtained information for subsequent reasoning according to the current reasoning state. In this paradigm, retrieval is not only used to supply background context, but also serves as an adaptive operation that can be invoked as the reasoning process unfolds. The model can refine search targets after resolving intermediate entities, and newly retrieved evidence can revise or extend subsequent reasoning. The design shifts the system from a static retrieve-then-generate pipeline to interleaved search and reasoning, producing search trajectories that contain search queries, retrieved evidence, intermediate reasoning, and candidate answers.

However, while deep search improves search depth and evidence coverage, it also introduces a new decision problem at the final selection stage. When multiple trajectories are generated for the same question, they may lead to different candidate answers and provide overlapping, redundant, incomplete, or even conflicting evidence support. Even if the correct answer appears in some trajectories, final selection strategies that rely on answer strings, or direct comparison of lengthy raw trajectories may still fail to select it reliably. Such strategies either discard the evidence support behind each candidate answer or require the verifier to process noisy and redundant trajectory content. Therefore, after multi-trajectory search, final performance depends not only on whether the search process obtains useful evidence, but also on whether the evidence is organized at a granularity suitable for comparing candidate answers. We argue that final answer selection should compare the evidence support associated with each candidate answer, and we formalize the process as answer-level evidence compression.

Based on this view, we propose STEC, an answer selection framework built on multi-trajectory outputs. Given the outputs for a question, STEC first organizes trajectories by candidate answer and constructs a candidate-specific evidence representation for each answer group. An evidence-guided verifier then compares these evidence representations and selects the final answer from the existing candidate answers. By converting the verification target from raw trajectories to candidate-specific evidence representations, STEC preserves the coverage benefit of multi-trajectory search while reducing the influence of redundant trajectories and noisy reasoning on final answer selection.

In summary, our contributions are threefold: (1) we identify final answer selection after multi-trajectory search as a candidate-centered evidence comparison problem; (2) we propose STEC, an evidence compression framework that integrates Answer-Level Evidence Compression and Evidence-Guided Answer Verification for candidate-specific evidence organization and evidence-based final selection; and (3) we evaluate STEC on four open-domain multi-hop question answering datasets, showing that it outperforms representative baselines and that ablation studies and quantitative analyses support the role of answer-level evidence compression.

\section{Related Work}

\subsection{Deep Search}

Deep search extends retrieval-augmented reasoning from single-round knowledge augmentation to iterative information seeking. Traditional RAG~\cite{ref1} mitigates knowledge limitations in language models on knowledge-intensive tasks by incorporating external non-parametric knowledge, but it typically follows a retrieve-then-generate pipeline. ReAct~\cite{ref2}, IRCoT~\cite{ref3}, and Self-RAG~\cite{ref4} further interleave reasoning with actions, retrieval, or self-reflection, allowing models to use external information dynamically during inference. Recent studies extend the paradigm to deep search with large reasoning models. Search-o1~\cite{ref5} introduces agentic search into long-chain reasoning, while RAG-Star~\cite{ref6} uses retrieval for verification and refinement in deliberative reasoning. WebThinker~\cite{ref7}, DeepResearcher~\cite{ref8}, and HierSearch~\cite{ref9} further study deep search in Web exploration, real-world research environments, and hierarchical search over local and Web knowledge sources. In parallel, Search-R1~\cite{ref10}, ReSearch~\cite{ref11}, R1-Searcher~\cite{ref12}, and ZeroSearch~\cite{ref13} improve models' search interaction capabilities through reinforcement learning or simulated search. SmartSearch~\cite{ref14}, Thinker~\cite{ref15}, and EvolveSearch~\cite{ref16} improve deep search from the perspectives of query refinement, hierarchical reasoning, and self-evolution, respectively. Overall, the line of work mainly focuses on information acquisition, search decision making, and reasoning organization during the search process. In contrast, STEC focuses on evidence consolidation and constructs candidate-specific evidence representations for final answer selection.

\subsection{Verifier-Based Test-Time Scaling}

Verifier-based test-time scaling allocates additional inference-time computation to verification, using verifier feedback to assess candidate solutions and support comparison, refinement, and selection. Existing work can be organized into training-based and training-free verifiers. Training-based methods are commonly instantiated as outcome reward models (ORMs) and process reward models (PRMs). At the outcome level, ORMs learn to rank complete solutions from outcome supervision and have been applied to logical reasoning~\cite{ref17} and Chain-of-Thought solution ranking~\cite{ref18}. At a finer granularity, PRMs provide feedback for intermediate reasoning states, enabling dense process-based verification and progress-aware verification~\cite{ref19}, as well as generative process judgment~\cite{ref20}. Although training-based verifiers provide explicit reward signals, they introduce additional costs in supervision collection and reward-model optimization. Training-free verification reduces these costs by repurposing existing models for inference-time judgment without task-specific verifier training. Existing training-free methods include Verification@$K$~\cite{ref21}, which scales verification over sampled responses; SETS~\cite{ref30}, which combines self-verification with self-correction; Multi-Agent Verification~\cite{ref31}, which aggregates multiple off-the-shelf verifier judgments; and Step-level Verifier-guided Hybrid Test-Time Scaling~\cite{ref32}, which moves verification to intermediate reasoning steps. Existing verifier-based test-time scaling methods mainly focus on verifier design and inference-time candidate selection. In contrast, STEC emphasizes evidence organization before verification, consolidating evidence from multiple search trajectories at the candidate level.

\section{Method}

\begin{figure}[t]
\centering
\includegraphics[width=\textwidth]{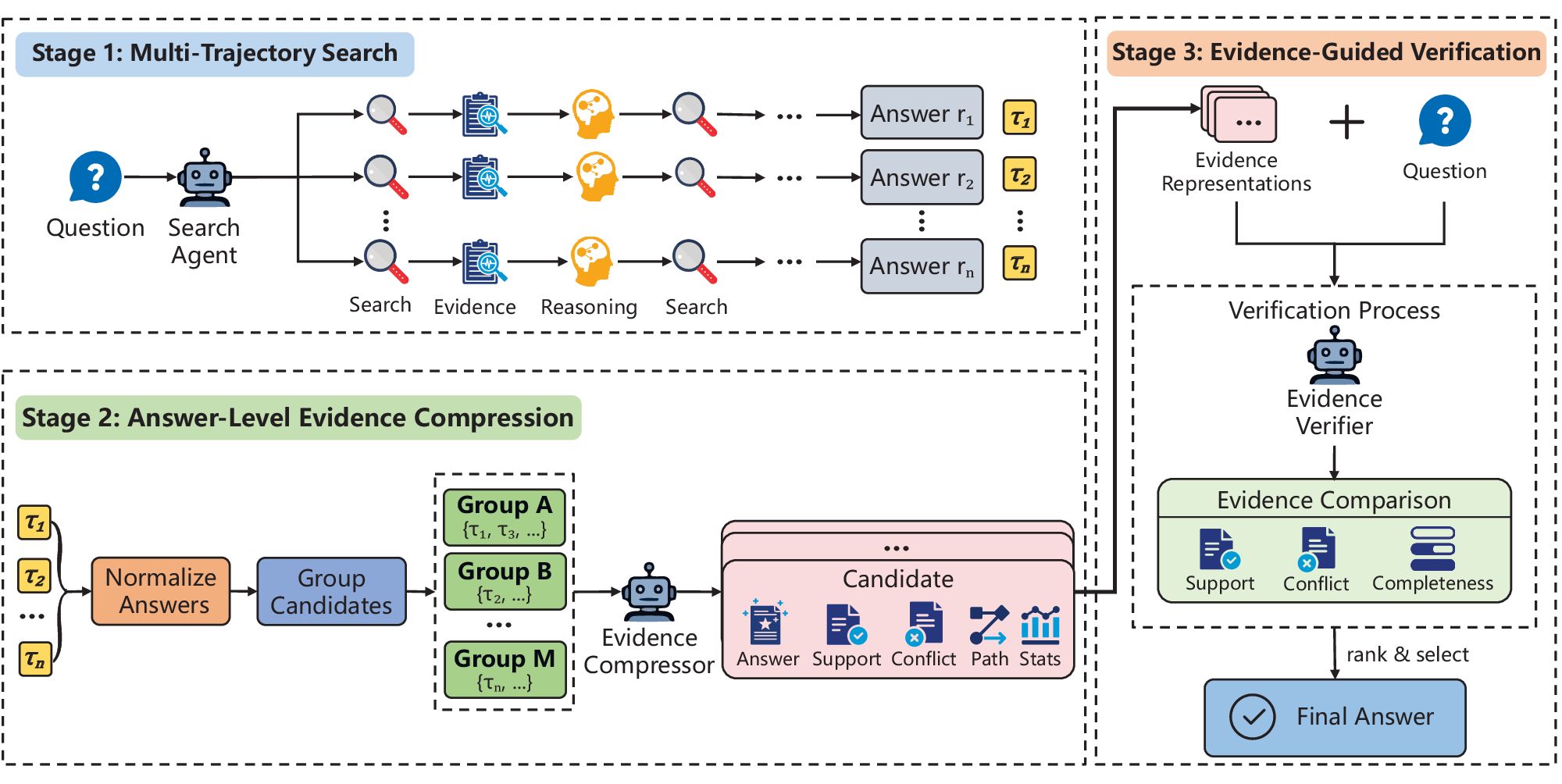}
\caption{Overview of STEC. Given a question, multi-trajectory search first produces search trajectories with candidate answers. STEC then groups trajectories by normalized answer identity, compresses each answer group into a candidate-specific evidence representation, and performs evidence-guided verification to select the final answer from the existing candidate set.}
\label{fig:stec_overview}
\end{figure}

We present STEC, a framework for final answer selection after multi-trajectory search. STEC consists of three stages: (1) Multi-Trajectory Search, which produces multiple trajectories for a question; (2) Answer-Level Evidence Compression, which groups trajectories by normalized answer identity and constructs candidate-specific evidence representations; and (3) Evidence-Guided Answer Verification, which compares these representations and selects the final answer from the existing candidate set. This design turns final selection into candidate-level evidence comparison rather than raw-trajectory comparison. Figure~\ref{fig:stec_overview} illustrates the overall workflow of STEC.

\begin{table}[t]
\caption{Notation summary.}
\label{tab:notation}
\centering
\small
\begin{tabular*}{\textwidth}{@{\extracolsep{\fill}}p{0.22\textwidth}p{0.72\textwidth}@{}}
\toprule
\textbf{Notation} & \textbf{Explanation} \\
\midrule
$q$ & input question \\
$N$ & number of search trajectories \\
$\mathcal{T}(q), \tau_i$ & trajectory set for $q$ and its $i$-th trajectory \\
$r_i, \tilde{a}_i$ & raw candidate answer and its normalized answer identity \\
$\mathcal{A}(q), M$ & candidate set and the number of unique candidates \\
$a$ & a candidate answer identity in $\mathcal{A}(q)$ \\
$G_a$ & trajectory group associated with candidate $a$ \\
$C_a$ & candidate-specific evidence representation \\
$V_a, E_a^{+}, E_a^{-}, P_a, \Gamma_a$ & answer variants, supporting evidence, conflicting evidence, reasoning path, and group statistics \\
$\mathcal{C}(q), \hat{a}$ & verification input and selected final answer identity \\
\bottomrule
\end{tabular*}
\end{table}

\subsection{Problem Definition and Our Approach}

We formulate final answer selection after multi-trajectory search as a candidate-centered evidence comparison problem. A search generator performs $N$ search rollouts for $q$ and produces a set of search trajectories:
\begin{equation}
\mathcal{T}(q)=\{\tau_1,\tau_2,\ldots,\tau_N\}.
\end{equation}
Each trajectory $\tau_i$ contains search interactions, retrieved evidence, intermediate reasoning steps, and a raw candidate answer $r_i$. The selection module takes $\mathcal{T}(q)$ as input and returns one final answer from the candidate answers produced by the search process. It does not generate a new answer outside the candidate set.

STEC addresses the problem by organizing search trajectories around candidate answers before verification. For each raw candidate answer $r_i$, STEC computes a normalized answer identity:
\begin{equation}
\tilde{a}_i=\operatorname{Norm}(r_i).
\end{equation}
The normalization function removes simple surface variations, such as case, punctuation, articles, and redundant whitespace. We use conservative normalization to reduce unsupported merging of answers with different meanings.

STEC defines the candidate set as the set of unique normalized answer identities:
\begin{equation}
\mathcal{A}(q)=\{a^{(1)},a^{(2)},\ldots,a^{(M)}\}
=\operatorname{Unique}\bigl(\{\tilde{a}_i\}_{i=1}^{N}\bigr),
\quad M\leq N.
\end{equation}
For each candidate $a\in\mathcal{A}(q)$, STEC constructs an answer group:
\begin{equation}
G_a=\{\tau_i\in\mathcal{T}(q)\mid \tilde{a}_i=a\}.
\end{equation}
The group $G_a$ contains all trajectories whose raw candidate answers map to the same normalized answer identity.

STEC then compresses each answer group into a candidate-specific evidence representation:
\begin{equation}
C_a=\operatorname{Compress}(q,a,G_a),
\quad a\in\mathcal{A}(q).
\end{equation}
All candidate-specific evidence representations form the verification input:
\begin{equation}
\mathcal{C}(q)=\{C_a\mid a\in\mathcal{A}(q)\}.
\end{equation}
The verifier compares these representations and selects the final answer:
\begin{equation}
\hat{a}=\operatorname{Verifier}\bigl(q,\mathcal{C}(q)\bigr),
\quad \hat{a}\in\mathcal{A}(q).
\end{equation}
The constraint keeps final selection grounded in the search outputs. After verification, STEC maps the selected normalized answer identity back to a representative raw answer string stored in the corresponding answer group.

\subsection{Multi-Trajectory Search}

Multi-Trajectory Search provides the trajectory set used by STEC for final selection. Given a question $q$, the search generator produces $N$ search trajectories. Each trajectory records one reasoning-and-retrieval process and ends with a raw candidate answer. These trajectories serve as the evidence acquisition output rather than the final decision.

Multi-trajectory search increases the chance of covering useful evidence and alternative reasoning paths. However, the resulting trajectories are not directly aligned for final comparison. Different trajectories may reach different candidate answers, and trajectories that lead to the same answer may contain overlapping, redundant, incomplete, or noisy evidence. Therefore, directly comparing raw trajectories can force the verifier to process heterogeneous and unstructured content. Conversely, comparing only answer strings discards the evidence basis behind each candidate. STEC therefore treats multi-trajectory search as an evidence acquisition stage and leaves candidate-centered evidence organization to the subsequent compression module.

\subsection{Answer-Level Evidence Compression}

Answer-Level Evidence Compression constructs a candidate-specific evidence representation for each normalized answer identity. Given the question $q$, a candidate answer identity $a$, and its answer group $G_a$, the module produces a compact representation $C_a$ that summarizes the evidence associated with candidate $a$. Different from generic trajectory summarization, the compression is performed separately for each candidate answer. Therefore, evidence from different answer groups is not mixed before verification.

Formally, the candidate-specific evidence representation is defined as
\begin{equation}
C_a=(V_a,E_a^{+},E_a^{-},P_a,\Gamma_a).
\end{equation}
Here, $V_a$ denotes the set of raw answer variants in $G_a$. $E_a^{+}$ denotes evidence units that support candidate answer $a$ under the constraints of question $q$. $E_a^{-}$ denotes evidence units that contradict $a$ or weaken its support. $P_a$ denotes a concise reasoning path from the supporting evidence to candidate answer $a$. $\Gamma_a$ denotes group-level statistics, such as the number of trajectories in $G_a$.

The compression process is performed for each answer group independently. First, the module collects the raw answer variants in $G_a$ and extracts compact evidence units from the retrieved content and intermediate reasoning steps associated with the same answer group. Each evidence unit corresponds to a short fact or passage segment from the original search trajectories, keeping the candidate-specific evidence representation grounded in the search outputs.

Second, the extracted evidence units are organized with respect to candidate answer $a$. Evidence that supports $a$ under the question constraints is placed in $E_a^{+}$, while evidence that contradicts $a$ or weakens its support is placed in $E_a^{-}$. Repeated evidence is merged to reduce redundancy. The module also records group-level statistics $\Gamma_a$, such as the number of trajectories associated with the candidate answer, to preserve trajectory-level support signals.

Finally, the module constructs a concise reasoning path $P_a$ from the supporting evidence. The path connects the question constraints to candidate answer $a$ without introducing new facts. As a result, each candidate answer is represented by the same set of fields, allowing Evidence-Guided Answer Verification to compare candidate-specific evidence representations rather than raw search trajectories.

The resulting representation makes different candidates comparable at the same granularity. Each candidate is represented by the same fields: answer variants, supporting evidence, conflicting evidence, reasoning path, and group statistics. The verifier can therefore compare structured candidate-level evidence rather than raw search trajectories with different lengths, formats, and noise levels. This design keeps final answer selection grounded in the original search trajectories while reducing the burden of processing redundant trajectory content.

\subsection{Evidence-Guided Answer Verification}

Evidence-Guided Answer Verification selects the final answer by comparing the candidate-specific evidence representations constructed in the previous stage. For a question $q$, the verification input is $\mathcal{C}(q)=\{C_a \mid a\in\mathcal{A}(q)\}$, where each $C_a$ corresponds to a normalized answer identity $a$. The verifier then selects one identity from the existing candidate set:
\begin{equation}
\hat{a}=\operatorname{Verifier}\bigl(q,\mathcal{C}(q)\bigr),
\quad \hat{a}\in\mathcal{A}(q).
\end{equation}
This formulation makes verification a constrained selection problem rather than open-ended answer generation. The final prediction is therefore grounded in the candidates produced by multi-trajectory search.

The comparison is guided by the structured fields in each candidate-specific representation. For a candidate $a$, the verifier considers whether the supporting evidence $E_a^{+}$ covers the entities, relations, and conditions required by the question, and whether the reasoning path $P_a$ connects the supporting evidence to the candidate answer through a coherent multi-hop reasoning process. It also considers evidence that weakens or contradicts the candidate, represented by $E_a^{-}$, since such evidence may indicate incomplete support or inconsistency with retrieved facts. Group-level statistics $\Gamma_a$ are used only as auxiliary signals for candidate comparison, rather than as sufficient evidence for correctness. The decision is therefore based on the overall evidence support associated with each candidate, rather than on answer strings or raw trajectories alone.

After verification, STEC maps the selected normalized answer identity $\hat{a}$ back to a representative raw answer string, which is retrieved from the corresponding answer group. This mapping keeps the final output consistent with the constrained selection result and prevents the verifier from introducing an answer outside the search outputs.

By verifying candidate-specific evidence representations, STEC avoids direct comparison over raw trajectories with different lengths, formats, and noise levels. The verifier compares candidates at the same granularity, which makes the final comparison aligned at the candidate level and less sensitive to redundant or noisy trajectory content.

\section{Experiments}

Experiments are conducted to evaluate STEC from four aspects, i.e., superiority, effectiveness, sensitivity, and generality, by answering the following four questions.

\begingroup
\renewcommand{\labelitemi}{$\bullet$}
\begin{itemize}
    \item \textbf{Q1: Superiority.} Does STEC outperform representative open-domain multi-hop QA baselines?

    \item \textbf{Q2: Effectiveness.} Does Answer-Level Evidence Compression improve final answer selection?

    \item \textbf{Q3: Sensitivity.} How does the performance fluctuation of STEC with different hyper-parameters?

    \item \textbf{Q4: Generality.} How does STEC perform across different model sizes and tuning types?
\end{itemize}
\endgroup

\subsection{Experimental Setup}

\paragraph{Datasets.}
We use four open-domain multi-hop QA datasets: HotpotQA~\cite{ref22}, 2WikiMultihopQA~\cite{ref23}, MuSiQue~\cite{ref24}, and Bamboogle~\cite{ref25}. HotpotQA provides multi-hop questions with sentence-level supporting facts; 2WikiMultihopQA combines structured and unstructured sources with reasoning-path evidence annotations; MuSiQue constructs multi-hop questions by composing connected single-hop questions; and Bamboogle contains handcrafted compositional questions for evaluating multi-step reasoning. Together, these datasets cover diverse retrieval and reasoning challenges and are suitable for evaluating final answer selection.

\begin{table}[t]
\caption{Dataset statistics. ``\#'' denotes the number of examples.}
\label{tab:dataset_stats}
\centering
\small
\begin{tabular*}{0.58\textwidth}{@{\extracolsep{\fill}}lcr@{}}
\toprule
Dataset & Split & \#Examples \\
\midrule
HotpotQA & dev & 7,405 \\
2WikiMultiHopQA & dev & 12,576 \\
MuSiQue & dev & 2,417 \\
Bamboogle & test & 125 \\
\bottomrule
\end{tabular*}
\end{table}

\paragraph{Evaluation Metric.}
We use Exact Match (EM) as the evaluation metric. EM measures whether the predicted answer exactly matches the reference answer after answer normalization. We report EM scores on each dataset and also report the macro-average EM score across the four datasets. The average EM provides an overall comparison across datasets while avoiding dominance by any single dataset.

\paragraph{Baselines.}
To evaluate the effectiveness of STEC, we compare it against the following baselines: (1) Inference without Retrieval, including Direct Inference and Chain-of-Thought (CoT) reasoning~\cite{ref26}; (2) Inference with Retrieval, including Retrieval-Augmented Generation (RAG)~\cite{ref1} and IRCoT~\cite{ref3}; (3) Search-Enhanced Reasoning, including Search-o1~\cite{ref5} and Search-R1~\cite{ref10}; and (4) Fine-Tuning-Based Methods, including supervised fine-tuning (SFT)~\cite{ref27}, R1-base~\cite{ref28}, R1-instruct~\cite{ref28}, and Rejection Sampling~\cite{ref29} with a search engine. The grouping compares STEC with representative open-domain QA methods across four experimental settings: inference without retrieval, inference with retrieval, search-enhanced reasoning, and fine-tuning-based methods.

\paragraph{Implementation.}
All methods are evaluated on the same full evaluation datasets using EM, with both per-dataset and average scores reported. To ensure comparability, we keep the retriever, knowledge corpus, number of retrieved documents, and pre-trained backbone model fixed whenever applicable. Unless otherwise specified, STEC uses \(N=8\) search trajectories generated by Search-R1 for each question. The evaluation design therefore focuses the comparison on the effectiveness of the proposed evidence-guided final selection framework.

\subsection{Main Results (RQ1)}

Table~\ref{tab} reports the main results for answering Q1. We compare
STEC with representative baselines from four families: inference without
retrieval, inference with retrieval, fine-tuning-based methods, and
search-enhanced reasoning. For a fair comparison, the main results use
Qwen2.5-7B as the backbone model and report both per-dataset EM and
macro-average EM across the four datasets.

\begin{table}[t]
\centering
\caption{Main results on four open-domain multi-hop QA datasets.}
\label{tab}
\small
\begin{tabular*}{\textwidth}{@{\extracolsep{\fill}}lccccc@{}}
\toprule
Method & HotpotQA & 2Wiki & MuSiQue & Bamboogle & Avg. \\
\midrule
Direct Inference & 0.183 & 0.250 & 0.031 & 0.120 & 0.146 \\
CoT & 0.092 & 0.111 & 0.022 & 0.232 & 0.114 \\
\midrule
RAG & 0.299 & 0.235 & 0.058 & 0.208 & 0.200 \\
IRCoT & 0.133 & 0.149 & 0.072 & 0.224 & 0.145 \\
\midrule
SFT & 0.217 & 0.259 & 0.066 & 0.112 & 0.164 \\
R1-base & 0.242 & 0.273 & 0.083 & 0.296 & 0.224 \\
R1-instruct & 0.237 & 0.292 & 0.072 & 0.293 & 0.224 \\
Rejection Sampling & 0.331 & 0.296 & 0.123 & 0.355 & 0.276 \\
\midrule
Search-o1 & 0.187 & 0.176 & 0.058 & 0.296 & 0.179 \\
Search-R1 & 0.389 & 0.230 & 0.162 & 0.424 & 0.301 \\
\midrule
\textbf{STEC} & \textbf{0.438} & \textbf{0.300} & \textbf{0.193} & \textbf{0.456} & \textbf{0.347} \\
\bottomrule
\end{tabular*}
\end{table}

STEC achieves the best overall performance, with an average EM of 0.347. It outperforms the strongest baseline in each method family. Compared with the best baselines from inference without retrieval, inference with retrieval, fine-tuning-based methods, and search-enhanced reasoning, STEC improves average EM by 0.201, 0.147, 0.071, and 0.046, respectively. These results show that the proposed evidence compression framework is effective across different open-domain QA settings.

At the dataset level, STEC achieves the best EM on all four benchmarks: 0.438 on HotpotQA, 0.300 on 2Wiki, 0.193 on MuSiQue, and 0.456 on Bamboogle. This dataset-level consistency shows that the overall advantage of STEC is not dominated by a single benchmark. Together with the average results, these observations answer Q1 positively: STEC achieves stronger open-domain multi-hop QA performance than representative baselines.

\subsection{Ablation Study (RQ2)}

To answer Q2, we compare STEC with Verifier Only. Verifier Only removes Answer-Level Evidence Compression while keeping the other settings unchanged. This controlled comparison examines whether candidate-level evidence organization benefits final answer selection. Table~\ref{tab:ablation} reports the ablation results.

\begin{table}[t]
\centering
\caption{Ablation study on Answer-Level Evidence Compression.}
\label{tab:ablation}
\small
\begin{tabular*}{\textwidth}{@{\extracolsep{\fill}}lccccc@{}}
\toprule
Method & HotpotQA & 2Wiki & MuSiQue & Bamboogle & Avg. \\
\midrule
Verifier Only & 0.431 & 0.291 & 0.194 & 0.440 & 0.339 \\
STEC & 0.438 & 0.300 & 0.193 & 0.456 & 0.347 \\
\bottomrule
\end{tabular*}
\end{table}

STEC improves the average EM from 0.339 to 0.347, showing that Answer-Level Evidence Compression provides a positive contribution to the overall performance. The gains are observed on HotpotQA, 2Wiki, and Bamboogle, while STEC remains comparable to Verifier Only on MuSiQue. This pattern suggests that candidate-level evidence organization is generally beneficial, especially when multiple trajectories provide complementary or partially redundant evidence. Instead of exposing the verifier to raw trajectory contents, Answer-Level Evidence Compression presents each candidate with a compact evidence representation, making the verification input more structured. These results provide empirical support for the effectiveness of Answer-Level Evidence Compression in answering Q2.

\subsection{Hyper-parameter Analysis on $N$ (RQ3)}

\begin{figure}[t]
\centering
\includegraphics[width=\textwidth]{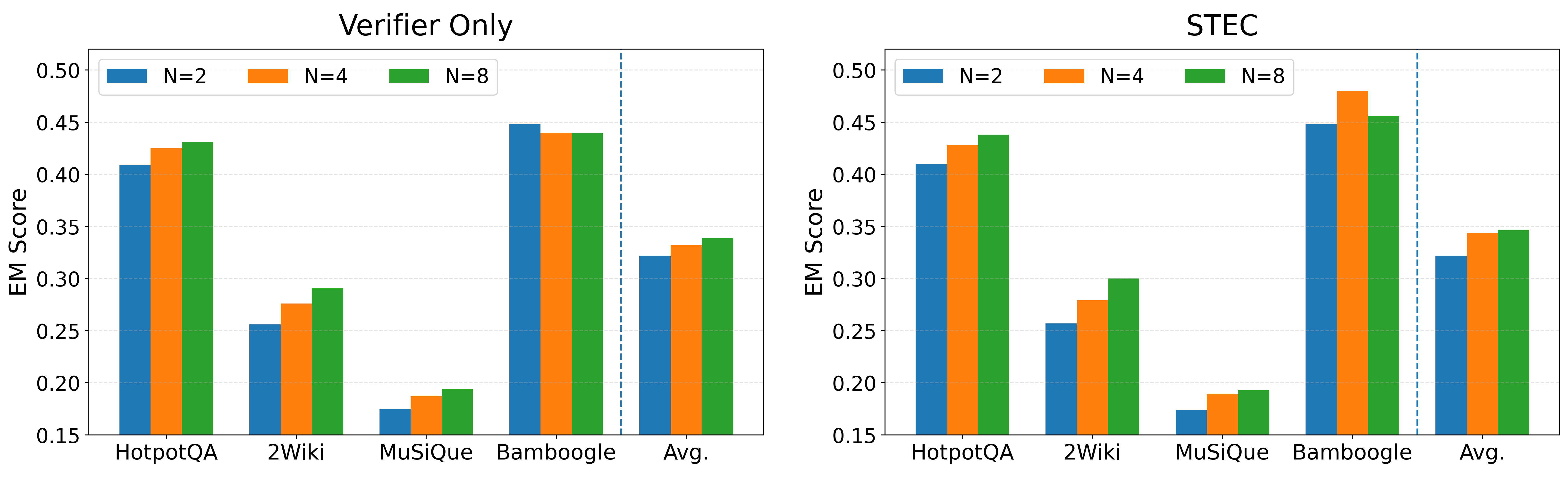}
\caption{Effect of the number of search trajectories. Left: EM scores of Verifier Only under different values of $N$. Right: EM scores of STEC under different values of $N$.}
\label{fig:trajectory_budget}
\end{figure}

To answer Q3, we vary the number of search trajectories from $N=2$ to $N=4$ and $N=8$. We report results for both Verifier Only and STEC to examine how the trajectory budget affects final performance. A larger $N$ can provide more candidate answers and evidence, but may also introduce redundancy and noise. Fig.~\ref{fig:trajectory_budget} presents the per-dataset results, and Table~\ref{tab:trajectory_budget} reports the average EM.

Increasing the trajectory budget $N$ generally improves average EM for both Verifier Only and STEC. For Verifier Only, the average EM increases from 0.322 at $N=2$ to 0.332 at $N=4$ and 0.339 at $N=8$, showing that additional trajectories provide useful information even without Answer-Level Evidence Compression. For STEC, the average EM increases from 0.322 to 0.344 and 0.347, indicating that STEC also benefits from a richer candidate and evidence pool. Comparing the two methods, STEC matches Verifier Only at $N=2$ and achieves higher average EM at $N=4$ and $N=8$. This comparison suggests that Answer-Level Evidence Compression remains effective under larger trajectory budgets.

The per-dataset results further show how STEC responds to different trajectory budgets. Performance on HotpotQA and 2Wiki improves as $N$ increases, whereas performance on MuSiQue remains stable. On Bamboogle, STEC achieves 0.480 EM
at $N=4$ and 0.456 EM at $N=8$, showing that a moderate trajectory budget can already provide strong evidence coverage. The per-dataset trend supports the effectiveness and stability of STEC under different values of $N$. Therefore, the results provide a positive answer to Q3: a larger trajectory budget generally benefits both Verifier Only and STEC, and Answer-Level Evidence Compression remains effective in multi-trajectory settings.

\begin{table}[t]
\centering
\caption{Average EM under different numbers of search trajectories.}
\label{tab:trajectory_budget}
\small
\begin{tabular*}{0.60\textwidth}{@{\extracolsep{\fill}}lccc@{}}
\toprule
Method & $N=2$ & $N=4$ & $N=8$ \\
\midrule
Verifier Only & 0.322 & 0.332 & 0.339 \\
STEC & 0.322 & 0.344 & 0.347 \\
\bottomrule
\end{tabular*}
\end{table}

\subsection{Model Analysis (RQ4)}
To answer Q4, we examine the performance of STEC under different model configurations. We compare STEC with Verifier Only under 3B Base, 3B Instruct, 7B Base, and 7B Instruct settings. In each configuration, trajectory generation
and final verification use the same model, and the remaining settings are kept consistent. Verifier Only removes Answer-Level Evidence Compression, so the comparison with STEC also reflects the contribution of the compression module under different model sizes and tuning types. To control the computational cost, we use 1,000 examples from each of HotpotQA, 2Wiki, and MuSiQue, together with the full Bamboogle dataset. Table~\ref{tab:model_config} reports the results.

\begin{table}[t]
\centering
\caption{Model configuration analysis under different model sizes and tuning types.}
\label{tab:model_config}
\small
\begin{tabular*}{\textwidth}{@{\extracolsep{\fill}}llccccc@{}}
\toprule
Model & Method & HotpotQA & 2Wiki & MuSiQue & Bamboogle & Avg. \\
\midrule
3B Base & Verifier Only & 0.222 & 0.219 & 0.046 & 0.128 & 0.154 \\
3B Base & STEC & 0.286 & 0.242 & 0.073 & 0.176 & 0.194 \\
\midrule
3B Instruct & Verifier Only & 0.287 & 0.240 & 0.091 & 0.208 & 0.207 \\
3B Instruct & STEC & 0.361 & 0.317 & 0.112 & 0.320 & 0.278 \\
\midrule
7B Base & Verifier Only & 0.446 & 0.282 & 0.177 & 0.440 & 0.336 \\
7B Base & STEC & 0.451 & 0.284 & 0.182 & 0.456 & 0.343 \\
\midrule
7B Instruct & Verifier Only & 0.422 & 0.367 & 0.159 & 0.456 & 0.351 \\
7B Instruct & STEC & 0.450 & 0.383 & 0.177 & 0.432 & 0.361 \\
\bottomrule
\end{tabular*}
\end{table}

Both Verifier Only and STEC are affected by model configuration. For Verifier Only, the Instruct variant achieves higher average EM than the corresponding Base variant under both 3B and 7B settings. The same trend is observed for STEC. This is reasonable because instruction-tuned models are better aligned with task instructions and output requirements. In addition, the 7B models generally outperform the 3B models, showing that model capacity also affects final verification performance.

The comparison between STEC and Verifier Only further shows the contribution of Answer-Level Evidence Compression. Across all four model configurations, STEC achieves higher average EM than Verifier Only. The gains are more visible under the 3B configurations, suggesting that compact candidate-level evidence is especially helpful when the verifier has more limited capacity. The consistent improvements across model configurations indicate that Answer-Level Evidence Compression provides complementary benefits to different backbone models.

The per-dataset results provide additional support. STEC improves most dataset-level results across the four configurations, with clear gains under the 3B Instruct setting on Bamboogle, 2Wiki, and HotpotQA. Under 7B Instruct, STEC decreases on Bamboogle but improves on the other three datasets, resulting in a higher average EM. Overall, the results provide a positive answer to Q4: STEC performs effectively across different model sizes and tuning types, and the comparison with Verifier Only shows that Answer-Level Evidence Compression remains beneficial under different model configurations.

\section{Conclusion}

In this paper, we propose STEC, an evidence compression framework for final answer selection after multi-trajectory search. STEC consists of two core components: (1) Answer-Level Evidence Compression, which groups search trajectories by normalized answer identity and constructs a candidate-specific evidence representation for each answer group; and (2) Evidence-Guided Answer Verification, which compares these representations and selects the final answer from the existing candidate set. By organizing evidence at the answer level, STEC preserves relevant candidate-specific evidence, thereby reducing the sensitivity of final verification to noisy trajectories. Experiments on four open-domain multi-hop QA datasets show that STEC achieves the best overall performance among the compared methods. Further analyses show that answer-level evidence compression helps organize multi-trajectory evidence and improves final answer selection.

%
%
%

%
%
%
%

\end{document}